\documentclass[11pt]{article}

\usepackage{acl}
\usepackage{times}
\usepackage{latexsym}
\usepackage[T1]{fontenc}
\usepackage[utf8]{inputenc}
\usepackage{microtype}
\usepackage{graphicx}
\usepackage{booktabs}
\usepackage{algorithm}
\usepackage{algpseudocode}
\usepackage{multirow}
\usepackage{float}
\usepackage{amsmath}
\usepackage{amssymb}
\usepackage{amsfonts}
\usepackage{enumitem}

\title{\textbf{SeDT: Sentence-Transformer Decision-Transformer\\
Conditioning for Multi-Turn Conversation Reliability}}

\author{
  Ramakrishna Vamsi Setti\textsuperscript{1} \quad
  Jagadeesh Rachapudi\textsuperscript{2} \quad
  Sachin Chaudhary\textsuperscript{3} \\
  \textbf{Praful Hambarde\textsuperscript{2}} \quad
  \textbf{Amit Shukla\textsuperscript{2}} \\[5pt]
  \textsuperscript{1}Independent Researcher \quad
  \textsuperscript{2}Drone Lab, IIT Mandi, \quad
  \textsuperscript{3}UPES, Dehradun \\[2pt]
  {\small \texttt{vamsisetti007@gmail.com, s23096@students.iitmandi.ac.in}}\\  
  {\small \texttt{sachin.chaudhary@ddn.upes.ac.in}}\\
  {\small \texttt{\{praful, amitshukla\}@iitmandi.ac.in}}
}

\begin{document}
\maketitle

\begin{abstract}
Large language models (LLMs) achieve impressive performance when a task is fully specified in a single turn, yet the same models lose up to $39\%$ of that performance when the identical task is revealed incrementally across multiple turns, a phenomenon documented at scale as \emph{Lost in Conversation}. Crucially, this collapse is almost entirely a \emph{reliability} failure; the best case, the aptitude only falls $16\%$, while the unreliability more than doubles ($+112\%$). We argue that the root cause is structural, a flat conversation history assigns equal implicit weight to every prior turn, giving the model no signal to distinguish a critical constraint from incidental dialog. We present \textbf{SeDT} \textbf{Se}ntence-transformer \textbf{D}ecision-\textbf{T}ransformer, a training-free inference-time method that resolves this by importing return-to-go conditioning from offline reinforcement learning. SeDT annotates each conversation shard with a cumulative relevance score derived from three complementary semantic, lexical, and positional signals and presents the full annotated history to the model at the final turn, without weight changes, without training data, and without discarding context. Evaluated on the Lost-in-Conversation benchmark in three LLMs and three generation tasks, SeDT outperforms the sharded baseline in all nine model-task combinations, with gains up to $+37.7$\% in mean performance $\bar{P}$ and simultaneous reductions in unreliability in seven of the nine combinations. In short, telling the model which past turns matter is sufficient to substantially recover the performance lost in conversation.
\end{abstract}

\section{Introduction}
\label{sec:intro}

Ask a language model to write a function and it will succeed~\citep{rachapudi2026backflush,rachapudi2026bid,rachapudi2026repair}. Give it the same problem one constraint at a time, the function name first, the input format next, the edge cases last, and it will quietly fall apart.~\citet{laban2025llms} document this failure at scale, 15 state-of-the-art (SoTA) LLMs evaluated on six tasks over 200,000 simulated conversations suffer an average performance drop of $39\%$ when the same instruction is revealed across turns rather than all at once. This is not a contrived setting. Analysis of large scale real world LLM conversations confirms that multi-turn, underspecified interaction is the norm rather than the exception~\citep{zheng2023lmsys}, and that users, particularly novice ones, rarely specify all requirements upfront~\citep{herlihy2024overcoming}.

The cause is structural. A multi-turn conversation is concatenated into a flat context window in which every prior turn carries equal implicit weight. The model receives no signal about which turns specified critical constraints and which were conversational scaffolding. Transformer attention compounds this; models systematically neglect middle-context information even in single-turn settings~\citep{liu2024lost}, and multi-turn conversations stack this bias across turns the model cannot distinguish from noise. Four concrete failure modes follow directly, premature commitment to an answer before all constraints are revealed, over-reliance on incorrect intermediate responses, loss of middle-turn constraints in favor of the first and last turns, and verbose drift that introduces false assumptions~\citep{laban2025llms}.

The offline reinforcement learning community faced a structurally identical problem. An agent learning from a flat replay buffer has no signal about which transitions were valuable. The Decision Transformer~\citep{chen2021decision} resolved this by annotating each trajectory step with its return-to-go (RTG), telling the agent what mattered rather than leaving it to infer importance from a flat buffer. The same insight applies to multi-turn LLM inference, yet existing solutions fall short; finetuning requires curated multi-turn training data and weight modification making it impractical at inference time, and users reveal constraints incrementally not by choice but by necessity~\citep{herlihy2024overcoming}. We import this parallel directly into inference-time prompt construction. A multi-turn conversation is a trajectory; each \emph{shard}, an atomic piece of information revealed in a single turn, is a step, and the semantic relevance of that shard to the final output goal is the reward. Just as the Decision Transformer tells the agent which steps mattered, SeDT tells the model which turns matter.

To our knowledge, no prior work identifies equal implicit turn weighting in flat conversation contexts as the structural driver of multi-turn reliability collapse; prior work attributes the failure to intent misalignment~\citep{liu2026intent} or model unreliability~\citep{laban2025llms} rather than to the context representation itself.

We present \textbf{SeDT} (\textbf{Se}ntence-transformer \textbf{D}ecision-\textbf{T}ransformer conditioning), a  training-free inference-time method that resolves flat-context weighting by annotating each prior shard with a cumulative relevance score and presenting the full RTG-annotated history to the model at the final turn changing no weights, requiring no training data, and discarding no context. On the Lost-in-Conversation benchmark~\citep{laban2025llms}, SeDT consistently outperforms the sharded baseline across all three evaluated LLMs and tasks, with gains up to $+37.7$\% in mean performance $\bar{P}$ and simultaneous reductions in unreliability, confirming that the lost-in-conversation problem is at least partially one of context weighting, addressable at inference time without training. SeDT requires no task-specific data, no model modification, and at most one additional LLM call.

\paragraph{Contributions.}
Our main contributions are as follows:
\begin{itemize}[leftmargin=*, noitemsep]
  \item \textbf{Problem identification} Flat-context turn weighting identified as the structural root cause of multi-turn reliability collapse, with a formal parallel  to Decision Transformer RTG conditioning.
  \item \textbf{SeDT} A training-free inference-time method that requires zero model modification and zero training data.
  \item \textbf{Three-signal relevance} A semantic, lexical and positional relevance formulation that directly counteracts the four documented failure modes of multi-turn LLMs.
  \item \textbf{RTG-grounded self-correction} A two-guard correction mechanism that introduces zero hurt cases  while providing a conservative verification pathway.
\end{itemize}

\section{Background}
\label{sec:background}

\subsection{Lost in Conversation}

\citet{laban2025llms} document a systematic performance gap between single-turn, fully-specified interaction and multi-turn, underspecified interaction at scale. Their sharding framework decomposes a fully-specified instruction into atomic information shards revealed one per turn, enabling controlled comparison while holding task content constant. Running $n$ independent simulations per example at temperature $T{=}1.0$ yields three metrics: average performance $\bar{P}$ (mean score), aptitude $A_{90}$ (90th percentile, capturing best-case capability) and unreliability $U$ (the 90th$-$10th percentile gap, where lower is better). The central finding is that the lost-in-conversation problem manifests itself primarily as an unreliability explosion ($+112\%$) rather than an aptitude
collapse ($-16\%$). Therefore, a reliable solution must reduce $U$ along with improving $\bar{P}$.

\subsection{Related Work}
\paragraph{Multi-turn evaluation}
A growing body of work evaluates LLMs in multi-turn settings through \emph{episodic} benchmark settings where each conversation turn introduces a self-contained subtask that can be evaluated independently, without requiring the model to fuse information accumulated across turns~\citep{zheng2023judging, bai2024mt, kwan2024mt, wang2023mint}. Although these benchmarks capture important capabilities such as refinement (iteratively improving a response based on user feedback) and tool use (calling external APIs or executing code across turns), they do not require the model to fuse underspecified information accumulated across turns and, as a result, systematically overestimate multi-turn performance~\citep{laban2025llms}.

\paragraph{Attention bias and long-context failures}
Transformer models exhibit a well-documented U-shaped attention bias, disproportionately attending to tokens at the beginning and end of long contexts while neglecting middle content~\citep{liu2024lost}. This phenomenon extends naturally to multi-turn conversation, where middle turns carrying critical constraints receive  insufficient attention~\citep{laban2025llms}. Static self-attention has been identified as the root cause of score dilution in long-context settings, motivating test-time approaches that address this at the query level~\citep{bansal2025let}.

\paragraph{Context management and recapitulation }
Beyond evaluation, a separate line of work addresses the multi-turn problem by modifying how conversation history is presented to the model. One family of approaches uses recapitulation, a strategy in which prior user turns are restated verbatim in the current context to ensure that the model has access to all previous constraints. Standard recapitulation appends all prior shards as a restatement at the final turn, while snowball recapitulation grows this restatement cumulatively at every turn~\citep{laban2025llms}. A subtler alternative is starting a new conversation for each sub-task, which empirically improves performance~\citep{laban2025llms} by resetting the flat context, but this discards conversational history entirely and is unavailable when constraints arrive incrementally from a real user. 
Finetuning resolves flat weighting by updating model weights on curated multi-turn data, but requires task-specific corpora, retraining for every new model or domain, and is unavailable at inference time.

\paragraph{Intent alignment and instruction following}
When faced with incomplete or ambiguous questions, LLMs exhibit systematic response patterns. Some models hedge, producing vague or noncommittal answers that avoid committing to an incomplete specification. Others issue clarification requests, asking the user to provide missing information before proceeding. A third pattern is the premature direct response, where the model assumes the most likely interpretation and responds immediately, often incorrectly~\citep{herlihy2024overcoming}. All three patterns reflect the same underlying problem; the model cannot determine user intent from an incomplete context. Improving multi-turn instruction following has been pursued through demonstration-guided training~\citep{sun2024parrot} and through architectural frameworks that decouple intent inference from task execution by reconstructing a single-turn instruction from conversation history~\citep{liu2026intent}.

\paragraph{Return-to-go conditioning}
Sequence modeling approaches to offline reinforcement learning have demonstrated that conditioning on desired outcomes, rather than learning value functions, produces robust and controllable agent behavior~\citep{chen2021decision}. Dense sentence representations optimized for semantic similarity have since enabled goal-conditioned reasoning in language settings~\citep{reimers2019sentence, song2020mpnet}, providing the computational foundation for translating outcome-conditioned sequence modeling into prompt-level context management. SeDT builds on both lines of work.
\section{Method: SeDT}
\label{sec:method}

\subsection{The Formal Parallel}

\begin{figure}[t]
\centering
\includegraphics[width=\columnwidth]{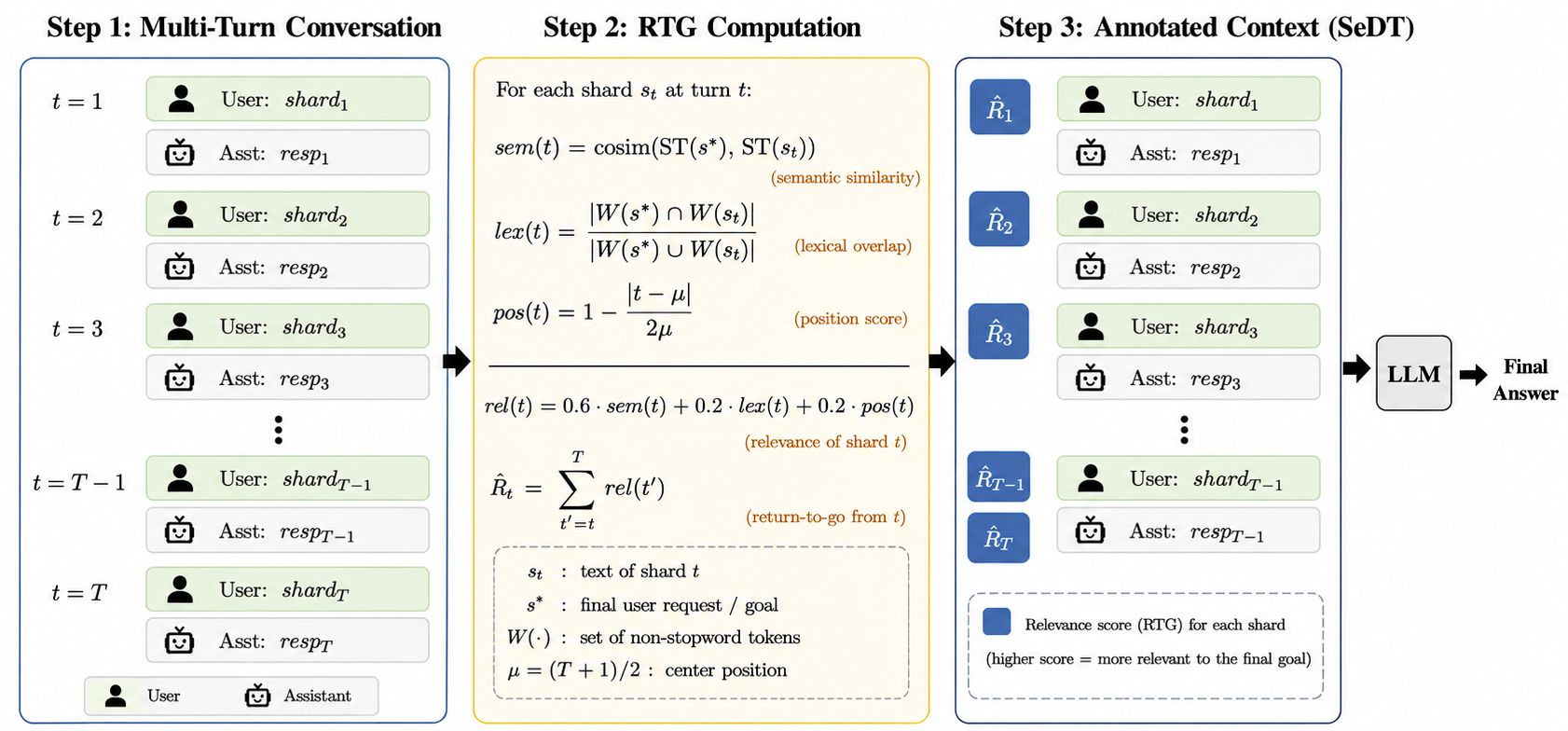}
\caption{Overview of the SeDT pipeline.}
\label{fig:method}
\end{figure}

The structural analogy between the Decision Transformer~\citep{chen2021decision} and SeDT is precise. In the Decision Transformer, a trajectory is a sequence of interactions between an agent and its environment recorded as a series of steps. At each step $t$, the agent observes a state $s_t$ describing the current situation, takes an action $a_t$, and receives a signal $\hat{R}_t$ called the return-to-go, which represents the total reward the agent can still accumulate from that step onward to the end of the trajectory. By conditioning on these return-to-go values, the Decision Transformer teaches the agent to associate high-value steps with high future reward, producing the full trajectory $[(\hat{R}_1, s_1, a_1), \ldots, (\hat{R}_T, s_T, a_T)]$ as input to generate the next action.
In SeDT, we draw a direct parallel to multi-turn conversation. The conversation history plays the role of the trajectory, each shard revealed in a single turn plays the role of a step, and the semantic relevance of that shard to the final output goal plays the role of reward. Just as the Decision Transformer annotates each step with how much future reward it is worth, SeDT annotates each shard with how much goal-relevant information is still to come from that turn onward, which we also call the return-to-go of that shard. The conversation history thus becomes $[(\hat{R}_1, \text{shard}_1), \ldots, (\hat{R}_T, \text{shard}_T)]$, conditioning the final answer on return-to-go values:
\begin{align*}
\text{DT:}\quad   & [\hat{R}_1, s_1, a_1]\;[\hat{R}_2, s_2, a_2]\;\cdots \to a_T \\
\text{SeDT:}\quad & [\hat{R}_1, \text{shard}_1]\;[\hat{R}_2, \text{shard}_2]\;\cdots \to \hat{y}_T,
\end{align*}
where $\hat{R}_t = \sum_{t'=t}^{T-1} \mathrm{rel}(t')$ is the cumulative relevance from turn $t$ to the final turn $T$, and $T$ is the total number of shards in the conversation. In the Decision Transformer, a high $\hat{R}_t$ tells the agent that much future reward is still achievable from this step. In SeDT, a high $\hat{R}_t$ tells the model that much goal-relevant information is still to come from this turn, and the model should carefully attend to the constraints that follow it. The parallel is not merely metaphorical; it directly motivates every design decision that follows. Figure~\ref{fig:method} shows the overview of SeDT.

\subsection{Goal-Based Anchor and Shard Embedding}
The anchor $s^*$ is the expected output goal, which the model must produce at the final turn. We use task-typed anchors: \textit{``Calculate and give the final numerical answer''} (math), \textit{``Return all required function calls with correct parameters''} (actions), and \textit{``Write a complete Python function that solves the problem''} (code). For each shard $s_t$ with intermediate response $r_t$, we embed $\mathrm{ST}(s_t \| r_t)$ if $|r_t| \le 30$ characters and $\mathrm{ST}(s_t)$ otherwise, preventing the verbose intermediate output from distorting the relevance estimate.

\subsection{Three-Signal Relevance Scoring}

A shard may be semantically relevant to the goal without sharing surface keywords and vice versa. To capture complementary dimensions of relevance, we combine three signals:
\begin{equation}
\mathrm{rel}(t) = \alpha \cdot \mathrm{sem}(t) + \beta \cdot \mathrm{lex}(t) + \gamma \cdot \mathrm{pos}(t),
\end{equation}
with weights $\alpha{=}0.6$, $\beta{=}0.2$, $\gamma{=}0.2$, chosen to reflect the relative discriminative power of each signal as empirically validated through the analysis of the  signal contribution  in Section~\ref{sec:results}. Semantic similarity to the output goal carries the most weight because it captures goal relevance that surface overlap misses, while lexical and positional signals provide complementary corrections at lower weights.

\textbf{Signal 1 (Semantic)} Deep alignment between the shard and the goal anchor, computed as $\mathrm{sem}(t) = \mathrm{cosim}(\mathrm{embed}(s^*), \mathrm{embed}_t)$, where $\mathrm{embed}(\cdot)$ denotes dense vector representations produced by the all-mpnet-base-v2 sentence transformer~\citep{reimers2019sentence} and $\mathrm{cosim}$ denotes cosine similarity between two vectors.

\textbf{Signal 2 (Lexical)} Surface keyword overlap between the goal anchor and the shard, computed as $\mathrm{lex}(t) = |W(s^*) \cap W(s_t)| / |W(s^*) \cup W(s_t)|$, where $W(s)$ denotes the set of unique words in the string $s$, $\cap$ is the intersection of sets and $\cup$ is the union of sets. This is the Jaccard similarity between the word sets of the anchor and the shard, capturing surface-level overlap that semantic embeddings may miss when domain-specific keywords appear verbatim across both.

\textbf{Signal 3 (Positional)} An inverted-U boost that counteracts both the suffix-sum bias of the RTG computation and the U-shaped positional attention of transformers~\citep{liu2024lost, vaswani2017attention}, calculated as $\mathrm{pos}(t) = 1 - |t - \mu| / (2\mu)$, where $t \in \{0, \ldots, T{-}1\}$ is the zero-indexed position of the current shard, $T$ is the total number of shards in the conversation, and $\mu = (T{-}1)/2$ is the midpoint of the shard indices. The intuition is straightforward; the formula assigns the highest score of $1.0$ to the middle turn and decreases symmetrically towards both first and last turns. This directly counteracts the transformer's natural tendency to over-attend to the beginning and end of the context while neglecting the middle, ensuring that constraints revealed in the middle of the conversation receive a relevance boost proportional to how far they are from the endpoints. This signal directly addresses the loss-of-middle-turns failure mode documented in~\citet{laban2025llms}.
A \emph{minimum guarantee} for the last shard ensures that the final shard, which by the construction of RTG accumulates the lowest cumulative score, receives an RTG of at least the mean RTG across all shards:
$\hat{R}_{T{-}1} \leftarrow \max(\hat{R}_{T{-}1},\; \frac{1}{T}\sum_t \hat{R}_t)$.

Table~\ref{tab:rtg_example} illustrates the full RTG annotation on a real GSM8K example. Shards above the mean relevance are marked \textbf{CONFIRMED}; below-average shards are marked \textbf{UNCERTAIN}. In particular, shard $t{=}4$ has the lowest raw relevance ($\mathrm{rel}{=}0.12$), yet the last-shard guarantee increases its RTG from $0.12$ to $0.61$, preventing the model from discarding the final constraint.

\begin{table*}[t]
\centering
\small
\setlength{\tabcolsep}{4pt}
\begin{tabular}{cl|rrrrrrl}
\toprule
$t$ & Shard & sem & lex & pos & rel & RTG & Label \\
\midrule
0 & What \% of students in hip-hop dance?    & .08 & .00 & .50 & .15 & 1.03 & UNCERTAIN \\
1 & There are 20 students in the dance class & .13 & .07 & .75 & .24 & 0.88 & CONFIRMED \\
2 & 20\% of the students chose contemporary  & .11 & .08 & 1.00 & .28 & 0.64 & CONFIRMED \\
3 & 25\% of the remaining opted for jazz     & .13 & .07 & .75 & .24 & 0.36 & CONFIRMED \\
4 & The rest are in the hip-hop dance group  & $-$.01 & .13 & .50 & .12 & 0.61$^\dagger$ & UNCERTAIN \\
\bottomrule
\end{tabular}
\caption{RTG annotation for a real GSM8K example with anchor \textit{``Calculate and give the final numerical answer''}. $^\dagger$Raised by the last-shard minimum guarantee from raw $0.12$.}
\label{tab:rtg_example}
\end{table*}

\subsection{RTG-Conditioned Prompt and Confidence Labels}

In the final turn, SeDT presents the complete history annotated with RTG and augments the system prompt with confidence labels: shards with $\mathrm{rel}(t) \geq \bar{\rho}$ are marked \textbf{CONFIRMED}; below-average
shards are marked \textbf{UNCERTAIN}. This gives the model both a quantitative relevance signal and a categorical one, reinforcing the constraints to prioritize and directly targeting all four failure modes of~\citet{laban2025llms}. \textbf{Premature commitment} is addressed by RTG reorienting attention toward the full goal-weighted history rather than the earliest shard. \textbf{Overreliance on incorrect intermediates} is reduced by UNCERTAIN labeling, which reduces the influence of low-relevance prior responses. \textbf{Loss of middle turns} is countered by Signal~3, which explicitly increases the relevance of middle turns. \textbf{Verbose drift} is prevented by excluding intermediate responses longer than 30 characters from embedding, preventing noisy outputs from inflating relevance scores.

\subsection{RTG-Guided Self-Correction}
\label{sec:selfcorrection}

When the final-turn RTG signal indicates a weak conversation, SeDT triggers a verification step. Na\"ive triggers such as $\hat{R}_{\text{last}} \le \bar{R}$ are unsuitable because RTG is a suffix sum, making this condition trivially always true.
Instead, we define a \emph{weakness score}:
\begin{equation}
W = (-z_{\text{last}}) \times u_r,
\end{equation}
where $z_{\text{last}} = (\mathrm{rel}(T{-}1) - \bar{\rho}) / \sigma_{\rho}$ is the normalized relevance of the final shard and $u_r$ is the fraction of UNCERTAIN shards. The adaptive threshold $\tau = 0.5 \times (1 - \bar{\rho})$ is reduced when the global conversation is weak globally, allowing for a more aggressive correction. Self-correction occurs when $W > \tau$.

The verifier's response replaces the initial answer only via \emph{RTG-grounded acceptance}: the replacement is accepted if and only if $a_1 \neq a_v$ and $\mathrm{cov}(v, \mathcal{C}) > \mathrm{cov}(r_1, \mathcal{C})$, where $\mathcal{C}$ is the CONFIRMED shard set and coverage is defined as
\begin{equation}
\mathrm{cov}(y, \mathcal{C}) = \frac{1}{|\mathcal{C}|} 
\sum_{s \in \mathcal{C}} \mathbf{1}\!\left[\exists\, w \in 
W_{>3}(s) : w \in W(y)\right],
\end{equation}
the fraction of confirmed shards for which at least one content word (longer than three characters) appears in the response $y$. This guard prevents a stochastic verifier from overwriting a correct answer simply by disagreeing.

\begin{algorithm}[t]
\caption{SeDT: RTG-Conditioned Multi-Turn Inference}
\label{alg:sedt}
\footnotesize
\begin{algorithmic}[1]
\Require Shards $\{s_1,\dots,s_T\}$, intermediate responses $\{r_1,\dots,r_{T-1}\}$,
         goal anchor $s^*$, LLM $\mathcal{M}$,
         weights $\alpha{=}0.6$, $\beta{=}0.2$, $\gamma{=}0.2$
\Ensure  Final answer $\hat{y}_T$
\For{$t = 0$ \textbf{to} $T{-}1$}
    \State $\text{text}_t \leftarrow s_t \| r_t$ \textbf{ if } $|r_t| \leq 30$ \textbf{ else } $s_t$
    \State $\mathbf{e}_t \leftarrow \mathrm{ST}(\text{text}_t)$
\EndFor
\State $\mathbf{e}^* \leftarrow \mathrm{ST}(s^*)$
\For{$t = 0$ \textbf{to} $T{-}1$}
    \State $\mathrm{sem}(t) \leftarrow \mathrm{cosim}(\mathbf{e}^*, \mathbf{e}_t)$
    \State $\mathrm{lex}(t) \leftarrow |W(s^*) \cap W(s_t)| \;/\; |W(s^*) \cup W(s_t)|$
    \State $\mu \leftarrow (T-1)/2$
    \State $\mathrm{pos}(t) \leftarrow 1 - |t - \mu|\,/\,(2\mu)$
    \State $\mathrm{rel}(t) \leftarrow \alpha\cdot\mathrm{sem}(t) + \beta\cdot\mathrm{lex}(t) + \gamma\cdot\mathrm{pos}(t)$
\EndFor
\For{$t = T{-}1$ \textbf{downto} $0$}
    \State $\hat{R}_t \leftarrow \sum_{t'=t}^{T{-}1} \mathrm{rel}(t')$
\EndFor
\State $\hat{R}_{T{-}1} \leftarrow \max(\hat{R}_{T{-}1},\; \tfrac{1}{T}\sum_t \hat{R}_t)$
\State $\bar{\rho} \leftarrow \tfrac{1}{T}\sum_t \mathrm{rel}(t)$
\For{$t = 0$ \textbf{to} $T{-}1$}
    \State $\ell_t \leftarrow \textbf{CONFIRMED}$ \textbf{ if } $\mathrm{rel}(t) \geq \bar{\rho}$ \textbf{ else } $\textbf{UNCERTAIN}$
\EndFor
\State Build annotated history $\mathcal{H} \leftarrow \{[\hat{R}_t,\, \ell_t,\, s_t,\, r_t]\}_{t=0}^{T{-}1}$
\State $\hat{y}_1 \leftarrow \mathcal{M}(\mathcal{H},\, s_T)$
\State $\sigma_\rho \leftarrow \mathrm{std}(\{\mathrm{rel}(t)\})$,\quad $u_r \leftarrow |\{t : \ell_t = \textbf{UNCERTAIN}\}| / T$
\State $z_{T{-}1} \leftarrow (\mathrm{rel}(T{-}1) - \bar{\rho})\,/\,\sigma_\rho$
\State $W \leftarrow (-z_{T{-}1}) \times u_r$
\State $\tau \leftarrow 0.5 \times (1 - \bar{\rho})$
\If{$W > \tau$}
    \State $\hat{y}_v \leftarrow \mathcal{M}(\mathcal{H},\, s_T,\, \text{verify})$
    \State $\mathcal{C} \leftarrow \{s_t : \ell_t = \textbf{CONFIRMED}\}$
    \If{$\hat{y}_v \neq \hat{y}_1$ \textbf{ and } $\mathrm{cov}(\hat{y}_v, \mathcal{C}) > \mathrm{cov}(\hat{y}_1, \mathcal{C})$}
        \State $\hat{y}_1 \leftarrow \hat{y}_v$
    \EndIf
\EndIf
\State \Return $\hat{y}_1$
\end{algorithmic}
\end{algorithm}

\section{Experimental Setup}
\label{sec:setup}

\paragraph{Dataset and tasks.}
We evaluate on the \texttt{microsoft/lost\_in\_conversation} HuggingFace dataset~\citep{laban2025llms}, using all available sharded examples: actions (105 instructions from the Berkeley Function Calling Leaderboard), math / GSM8K~\citep{cobbe2021training} (103 instructions), and code~\citep{chen2021evaluating, jain2024livecodebench}(100 instructions; 45 HumanEval + 55 LiveCodeBench medium). These three tasks represent the binary-correctness subset of the benchmark, where the evaluation is unambiguous: functional accuracy for code, exact match for API calls, and numerical match for math.

\paragraph{Models.}
We evaluate three LLMs spanning three families: GPT-4o-mini~\citep{achiam2023gpt} (OpenAI), Gemini 2.5 Flash~\citep{team2023gemini} (Google) and Llama 3.3-70B~\citep{grattafiori2024llama} (Meta), covering proprietary and open-weight models at different scales.

\paragraph{Evaluation protocol.}
Following~\citet{laban2025llms}, we perform $n{=}5$ independent simulations per example at temperature $T{=}1.0$, with \texttt{max\_tokens = 1000}. The Metrics $\bar{P}$, $A$, and $U$ are computed per sample from the $n$ run scores and averaged across the corpus. We terminate conversations early on a correct answer, preventing later incorrect turns from overwriting a correct intermediate; this is particularly consequential for code tasks, where partial specifications can yield valid partial solutions.

\paragraph{Sharded baseline.}
The Sharded baseline presents all prior shards as a plain multi-turn conversation with no relevance annotation. All other settings for LLM, the system prompt, and the scorer are identical to SeDT. Task-specific scorers follow the base paper exactly: flexible regex for math, AST-based evaluation for API calls, and test-case execution for code.

\section{Results}
\label{sec:results}

\begin{figure}[t]
\centering
\includegraphics[width=\columnwidth]{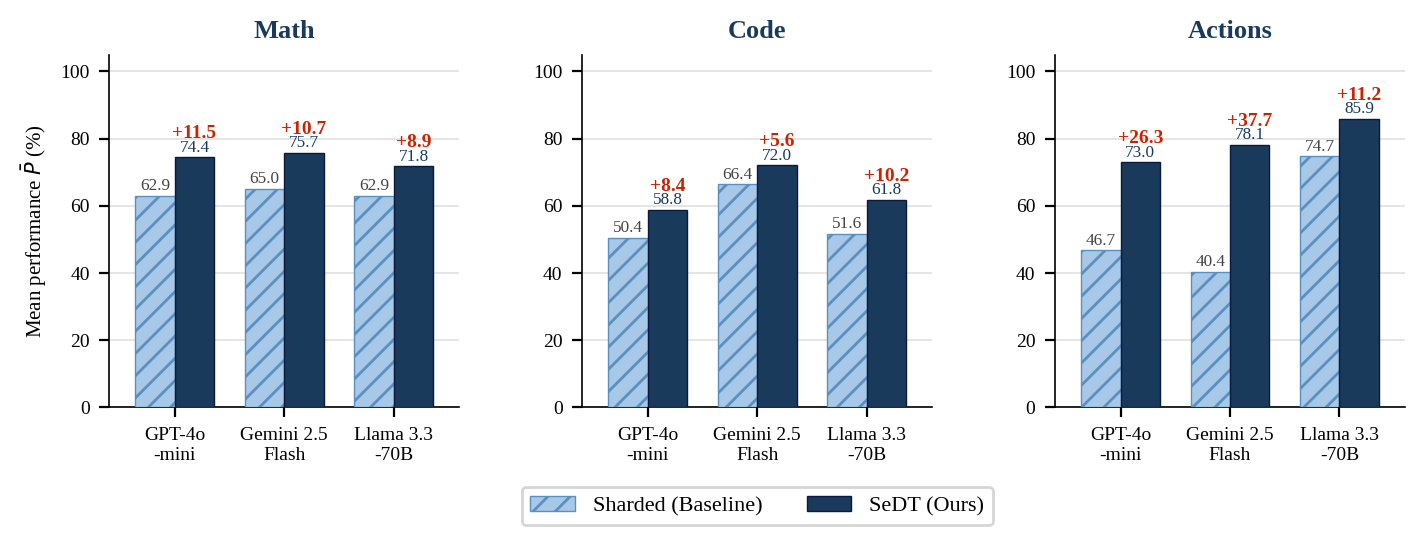}
\caption{Mean performance $\bar{P}$ (\%) for Sharded (hatched) and SeDT (solid) across three tasks and three model families. SeDT consistently outperforms the sharded baseline on all nine model-task combinations.}
\label{fig:results}
\end{figure}

\subsection{Main Results}

Table~\ref{tab:main} and Figure~\ref{fig:results} present the mean performance for all nine combinations of model tasks. SeDT consistently outperforms the flat-history sharded baseline on every model and every task.

In the actions task (105 examples), the gains range from $+11.2$ to $+37.7$. Gemini 2.5 Flash~\citep{team2023gemini} records the largest single gain ($+37.7$) and the highest average gain across all three tasks ($+18.0$ points), reaching $78.1$, more than halfway back to its single-turn ceiling of $88.4$. GPT-4o-mini~\citep{achiam2023gpt} and Llama 3.3-70B~\citep{grattafiori2024llama} also improve substantially, by $+26.3$ and $+11.2$, respectively.

In math (103 examples, GSM8K~\citep{cobbe2021training}), all three models improve consistently:
GPT-4o-mini~\citep{achiam2023gpt} gains $+11.5$, Gemini 2.5 Flash~\citep{team2023gemini} gains $+10.7$, and Llama 3.3-70B~\citep{grattafiori2024llama} gains $+8.9$. In code (100 examples), the gains hold for all models: $+8.4$, $+5.6$, and $+10.2$ for GPT-4o-mini~\citep{achiam2023gpt}, Gemini 2.5 Flash~\citep{team2023gemini} and Llama 3.3-70B~\citep{grattafiori2024llama}, respectively. The Code gains are smaller in absolute terms, reflecting the inherent difficulty of LiveCodeBench~\citep{jain2024livecodebench} medium problems, but SeDT improves consistently even in this challenging regime.

\begin{table*}[t]
\centering
\small
\setlength{\tabcolsep}{6pt}
\begin{tabular}{l|l|rrrr}
\toprule
\textbf{Model} & \textbf{Method}
  & \textbf{Actions} & \textbf{Math} & \textbf{Code}
  & \textbf{Average} \\
\midrule
\multirow{4}{*}{GPT-4o-mini~\citep{achiam2023gpt}}
  & Full$^\dagger$  & 94.1 & 88.1 & 75.9 & 86.0 \\
  & Sharded       & 46.7 & 62.9 & 50.4 & 53.3 \\
  & SeDT            & \textbf{73.0} & \textbf{74.4} & \textbf{58.8} & \textbf{68.7} \\
  & \textit{Gain}   & \textit{+26.3} & \textit{+11.5} & \textit{+8.4} & \textit{+15.4} \\
\midrule
\multirow{4}{*}{Gemini 2.5 Flash~\citep{team2023gemini}}
  & Full$^\dagger$  & 88.4 & 90.6 & 97.0 & 92.0 \\
  & Sharded       & 40.4 & 65.0 & 66.4 & 57.3 \\
  & SeDT            & \textbf{78.1} & \textbf{75.7} & \textbf{72.0} & \textbf{75.3} \\
  & \textit{Gain}   & \textit{+37.7} & \textit{+10.7} & \textit{+5.6} & \textit{+18.0} \\
\midrule
\multirow{4}{*}{Llama 3.3-70B~\citep{grattafiori2024llama}}
  & Full$^\dagger$  & 95.0 & 91.7 & 72.0 & 86.2 \\
  & Sharded       & 74.7 & 62.9 & 51.6 & 63.1 \\
  & SeDT            & \textbf{85.9} & \textbf{71.8} & \textbf{61.8} & \textbf{73.2} \\
  & \textit{Gain}   & \textit{+11.2} & \textit{+8.9} & \textit{+10.2} & \textit{+10.1} \\
\bottomrule
\end{tabular}
\caption{Mean performance $\bar{P}$ (\%) across three LLMs and three tasks.
Full$^\dagger$ is the single-turn upper bound from~\citet{laban2025llms};
Sharded is the flat-history multi-turn baseline; SeDT is our method.
Gain is the absolute improvement over Sharded.}
\label{tab:main}
\end{table*}

\subsection{Reliability Analysis}

Table~\ref{tab:reliability} presents aptitude $A$ and unreliability $U$ in all nine model-task combinations. SeDT improves $A$ on every combination, confirming that gains reflect genuine improvement in best-case capability rather than noise reduction, and reduces $U$ in seven of nine, with the strongest reductions in math (Llama 3.3-70B~\citep{grattafiori2024llama}, $\Delta U = -7.6$) and actions (Gemini 2.5 Flash~\citep{team2023gemini}, $\Delta U = -9.6$). The two exceptions Actions / Llama 3.3-70B~\citep{grattafiori2024llama} ($\Delta U = +0.2$) and Code / GPT-4o-mini~\citep{achiam2023gpt} ($\Delta U = +2.6$) both occur where the Sharded $U$ is already the lowest within its task, leaving limited room for further reduction — confirming that SeDT tightens score distributions precisely where the reliability improvement the lost-in-conversation literature calls for~\citep{laban2025llms} is most needed.

\begin{table*}[t]
\centering
\small
\setlength{\tabcolsep}{5pt}
\begin{tabular}{l|l|rrrrrr}
\toprule
 & & \multicolumn{2}{c}{\textbf{Sharded}} &
     \multicolumn{2}{c}{\textbf{SeDT} ours} &
     \multicolumn{2}{c}{\textbf{Delta}} \\
\cmidrule(lr){3-4} \cmidrule(lr){5-6} \cmidrule(lr){7-8}
\textbf{Task} & \textbf{Model}
  & $A$ & $U$
  & $A$ & $U$
  & $\Delta A$ & $\Delta U$ \\
\midrule
\multirow{3}{*}{Actions}
 & GPT-4o-mini~\citep{achiam2023gpt}
   & 54.3 & 14.5
   & \textbf{79.0} & \textbf{13.5}
   & $+24.7$ & $-1.0$ \\
 & Gemini 2.5 Flash~\citep{team2023gemini}
   & 56.4 & 30.7
   & \textbf{87.4} & \textbf{21.1}
   & $+31.0$ & $-9.6$ \\
 & Llama 3.3-70B~\citep{grattafiori2024llama}
   & 80.6 & 11.6
   & \textbf{91.0} & 11.8
   & $+10.4$ & $+0.2$ \\
\midrule
\multirow{3}{*}{Math}
 & GPT-4o-mini~\citep{achiam2023gpt}
   & 77.7 & 29.3
   & \textbf{88.9} & \textbf{28.9}
   & $+11.2$ & $-0.4$ \\
 & Gemini 2.5 Flash~\citep{team2023gemini}
   & 80.6 & 32.4
   & \textbf{85.8} & \textbf{23.5}
   & $+5.2$  & $-8.9$ \\
 & Llama 3.3-70B~\citep{grattafiori2024llama}
   & 84.3 & 44.3
   & \textbf{87.6} & \textbf{36.7}
   & $+3.3$  & $-7.6$ \\
\midrule
\multirow{3}{*}{Code}
 & GPT-4o-mini~\citep{achiam2023gpt}
   & 63.2 & 24.8
   & \textbf{72.6} & 27.4
   & $+9.4$  & $+2.6$ \\
 & Gemini 2.5 Flash~\citep{team2023gemini}
   & 80.2 & 27.6
   & \textbf{83.6} & \textbf{24.8}
   & $+3.4$  & $-2.8$ \\
 & Llama 3.3-70B~\citep{grattafiori2024llama}
   & 69.6 & 36.2
   & \textbf{76.8} & \textbf{31.6}
   & $+7.2$  & $-4.6$ \\
\bottomrule
\end{tabular}
\caption{Aptitude $A$ (90th percentile) and unreliability $U$ (90th$-$10th percentile gap, lower is better), computed per-example and averaged across the corpus following~\citet{laban2025llms}.}
\label{tab:reliability}
\end{table*}

\subsection{Signal Contribution Analysis}

To understand which of the three relevance signals drives performance, Table~\ref{tab:ablation} reports $\bar{P}$, $A$, and $U$ on a representative subset of 25 GSM8K~\citep{cobbe2021training} examples with GPT-4o-mini~\citep{achiam2023gpt}, ablating each signal in turn.

Each single-signal variant underperforms SeDT by $8.0$--$11.2$ points in $\bar{P}$, confirming that no individual signal is sufficient. The semantic signal alone ($\bar{P}{=}56.8$) performs worse despite receiving the highest weight in the full model, because cosine similarity without positional correction is vulnerable to suffix-sum bias. The position signal alone ($\bar{P}{=}60.0$) provides the strongest single-signal baseline, consistent with middle-turn neglect being a primary failure mode. Combining all three signals ($\bar{P}{=}68.0$) outperforms the best uniform weighting ($\bar{P}{=}63.2$) by $4.8$ points, confirming that $\alpha{=}0.6$, $\beta{=}0.2$, $\gamma{=}0.2$ reflects a meaningful prior on the importance of the signal rather than the choice of incidental hyperparameter.

To isolate whether the gains are due to the RTG prompt format or the calculated relevance signal, we include a Random RTG condition that replaces the computed scores with uniform random values in $[0,1]$ while keeping the prompt structure identical. Random RTG ($\bar{P}{=}60.8$) outperforms the sharded baseline, confirming that the instruction-tuned models respond to numerical prefixes from pretraining exposure~\citep{ouyang2022training, wei2022chain}. SeDT ($\bar{P}{=}68.0$) further outperforms Random RTG by $7.2$ points in $\bar{P}$ and by $13.6$ points in $U$ ($33.6 \to 20.0$), demonstrating that structured relevance adds substantial signal beyond the format alone.

\begin{table*}[t]
\centering
\small
\setlength{\tabcolsep}{6pt}
\begin{tabular}{l|cccccc}
\toprule
\textbf{Configuration} & $\alpha$ & $\beta$ & $\gamma$ & $\bar{P}$ & $A$ & $U$ \\
\midrule
Semantic only        & 1.00 & 0.00 & 0.00 & 56.8 & 70.4 & 29.6 \\
Lexical only         & 0.00 & 1.00 & 0.00 & 59.2 & 72.8 & 27.2 \\
Position only        & 0.00 & 0.00 & 1.00 & 60.0 & 73.6 & 27.2 \\
Uniform weights      & 0.33 & 0.33 & 0.34 & 63.2 & 75.2 & 24.0 \\
Random RTG           & 0.60 & 0.20 & 0.20 & 60.8 & 76.8 & 33.6 \\
\midrule
\textbf{SeDT}(ours) & 0.60 & 0.20 & 0.20 & \textbf{68.0} & \textbf{80.0} & \textbf{20.0} \\
\bottomrule
\end{tabular}
\caption{Signal contribution analysis on 25 GSM8K~\citep{cobbe2021training} examples with GPT-4o-mini~\citep{achiam2023gpt}.}
\label{tab:ablation}
\end{table*}

\subsection{Self-Correction Mechanism Analysis}
\label{sec:correction_analysis}

We evaluated the RTG-guided self-correction mechanism on 15 randomly drawn GSM8K~\citep{cobbe2021training} examples (75 runs total, \texttt{random.seed(42)}) with GPT-4o-mini, recording Guard~1 trigger and Guard~2 acceptance decisions at each run. The verifier triggers on $28.0\%$ of runs ($21/75$), indicating that $W > \tau$ fires selectively in genuinely uncertain conversations; in most triggered cases the verifier confirms the original answer without replacement, and when it disagrees Guard~2 blocks replacement unless the verifier demonstrates strictly higher confirmed-shard coverage. Across all 75 runs, \textbf{zero hurt cases} are recorded, confirming that the mechanism introduces no errors while providing a conservative verification pathway.

\subsection{Why RTG Conditioning Works Without Training}

A natural question is why a model responds meaningfully to return-to-go prefixes it was not explicitly trained on. Instruction-tuned LLMs are extensively exposed to relevance-annotated text during pretraining; code comments, document outlines, meeting agendas, and editorial markup all of which condition content on explicit importance signals, suggesting RTG-style numerical prefixes fall within their learned distribution~\citep{ouyang2022training, wei2022chain}. Additionally, the RTG-annotated history functions as an implicit in-context demonstration, the pattern $[\hat{R}_t, \ell_t, s_t]$ repeated through turns teaches the model that higher-scored shards carry more constraint weight, without any gradient signal~\citep{brown2020language}. Consistent with this, the two models with the highest gains, Gemini 2.5 Flash ($+18.0$) and GPT-4o-mini ($+15.4$), are the most capable instruction-tuned models in our evaluation, while Llama 3.3-70B records the lowest gain ($+10.1$).

\subsection{Comparison with Prior Approaches}

\paragraph{Recapitulation strategies}
\citet{laban2025llms} evaluate two recapitulation baselines on GPT-4o-mini~\citep{achiam2023gpt}; Recap, which includes a complete restatement of all prior shards as a final turn, and Snowball, which cumulatively restates shards at every turn. Both improve over the sharded baseline, yet fall short of the single-turn ceiling, with Recap achieving an average $\bar{P}$ of $66.5$ and Snowball $61.8$ across four tasks. The per-task breakdowns for these baselines are not publicly available, which excludes a direct per-task comparison. SeDT achieves $68.7$ on GPT-4o-mini across our three evaluated tasks, comparing favorably to both reported averages, while requiring no modification of the conversation structure and no assumption about which turn is final.

\paragraph{Intent reconstruction}
\citet{liu2026intent} propose a Mediator-Assistant framework that decouples intent inference from task execution, reconstructing ambiguous conversation history into a fully-specified single-turn instruction before passing it to the assistant. The Mediator requires task-specific historical interaction data and invokes 2--3 additional LLM calls per inference. On GPT-4o-mini~\citep{achiam2023gpt}, it achieves $\bar{P}$ of $85.7$, $77.7$ and $66.9$ on Actions, Math, and Code respectively~\citep{liu2026intent}. SeDT achieves $73.0$, $74.4$, and $58.8$, remaining competitive on Math while requiring zero task-specific data and at most one additional LLM call. The two approaches are complementary; the Mediator addresses intent alignment through experience-driven reconstruction, while SeDT addresses context weighting through return-to-go conditioning.

\section{Conclusion}
\label{sec:conclusion}


We identified flat-context turn weighting as a structural root cause of multi-turn reliability collapse and proposed \textbf{SeDT}, a training-free inference-time method importing return-to-go conditioning from offline reinforcement learning. The key insight is that conversation shards map to trajectory steps and semantic relevance serves as reward. SeDT annotates each shard with a cumulative relevance score from three complementary signals and presents the annotated history at the final turn, changing no weights and requiring no training data. Across three LLMs and three tasks, SeDT outperforms the sharded baseline in all nine combinations, improving mean performance while reducing unreliability in seven. The self-correction mechanism introduces zero hurt cases.

\section*{Limitations}

Several design decisions and resource constraints scope the current evaluation. We use $n{=}5$ runs per example rather than $n{=}10$ as in~\citet{laban2025llms}, following~\citet{liu2026intent} who also adopt $n{=}5$ and demonstrate stable estimates at this sample size; the $U$ metric may marginally underestimate the true variance. Due to computational constraints, we evaluated three of the six tasks in~\citet{laban2025llms}, as the remaining tasks (Database, Data-to-Text, Summary) require SQL execution infrastructure or long-context pipelines that exceed our resource budget. Goal anchors are task-typed natural language strings that describe the expected output, analogous to system prompt design and can be specified with minimal effort for any new task, though automatic anchor generation remains future work. The signal weights $\alpha{=}0.6$, $\beta{=}0.2$, $\gamma{=}0.2$ were validated on 25 GSM8K examples with GPT-4o-mini~\citep{achiam2023gpt}; however, consistent gains across all nine combinations of model-task with these fixed weights provide indirect evidence of robustness. A systematic weight sweep across all tasks and model families, complete six-task evaluation, and automatic anchor generation are left to future work.

\section*{E1 - AI Assistant Disclosure}
An AI writing assistant was used for paper-prose drafting, paragraph restructuring, and figure / table layout suggestions. The assistant was \textit{not} used to generate or fabricate experimental numbers, dataset records, annotator labels, or finding interpretations. All empirical results, dataset construction scripts, model selection decisions, are author-generated and have been independently verified against the locked release. The specific assistant and version will be named in the camera-ready; anonymising the vendor here preserves double-blind review.

\section*{Ethics Statement}
This work uses publicly available datasets and API-based language models. No human subjects were involved and no new data were collected. The proposed method is inference time only and introduces no additional risks beyond those of the underlying models.

\bibliography{main}

\end{document}